\documentclass[12pt, letterpaper]{extarticle}

\usepackage{subcaption}
\usepackage{fullpage}
\usepackage[switch]{lineno}
\usepackage{amsmath}
\usepackage{amssymb}
\usepackage{rotating}
\usepackage{array}
\usepackage{mathtools}
\usepackage[ruled]{algorithm2e}
\usepackage{algorithmic}
\usepackage{bm}
\usepackage{breqn}
\usepackage{comment}
\usepackage{enumitem}
\usepackage{graphics}
\usepackage{graphicx}
\usepackage{latexsym}
\usepackage{mathrsfs}
\usepackage{morefloats}
\usepackage{nicefrac}
\usepackage{authblk}
\usepackage{pifont}
\usepackage{nameref}

\usepackage[hyphens]{url}
\usepackage{hyperref}
\hypersetup{colorlinks=false,breaklinks=true}

\title{Neutralizing the Narrative: AI-Powered Debiasing of Online News Articles}

\author[1*]{Chen Wei Kuo}
\author[1*]{Kevin Chu}
\author[1*]{Nouar AlDahoul}
\author[1*]{Hazem Ibrahim}
\author[1+]{\\Talal Rahwan}
\author[1+]{Yasir Zaki}

\affil[1]{\normalsize New York University Abu Dhabi, UAE.}
\affil[*]{\footnotesize Authors contributed equally}
\affil[+]{\footnotesize Correspondence author. E-mail: \{talal.rahwan,yasir.zaki\}@nyu.edu}
\date{}

\begin{document}


\maketitle

\begin{abstract}
Bias in news reporting significantly impacts public perception, particularly regarding crime, politics, and societal issues. Traditional bias detection methods, predominantly reliant on human moderation and suffer from subjective interpretations and scalability constraints. Here, we introduce an AI-driven framework leveraging advanced large language models (LLMs), specifically GPT-4o, GPT-4o Mini, Gemini Pro, Gemini Flash, Llama 8B, and Llama 3B, to systematically identify and mitigate biases in news articles. To this end, we collect an extensive dataset consisting of over 30,000 crime-related articles from five politically diverse news sources spanning of a decade (2013–2023). Our approach employs a two-stage methodology: (1) Bias detection, where each LLM scores and justifies biased content at the paragraph level, validated through human evaluation for ground truth establishment, and (2) Iterative debiasing using GPT-4o Mini, verified by both automated reassessment and human reviewers. Empirical results indicate GPT-4o Mini's superior accuracy in bias detection and effectiveness in debiasing. Furthermore, our analysis reveals temporal and geographical variations in media bias correlating with socio-political dynamics and real-world events. This study contributes to scalable computational methodologies for bias mitigation, promoting fairness and accountability in news reporting.

\end{abstract}

\section{Introduction}
The prevalence of bias in news media significantly influences public perceptions, shaping public discourse on sensitive issues such as crime, politics, and social justice \cite{trhlik_quantifying_2024}. This bias is often topic-specific, appearing in coverage related to issues like COVID-19, immigration, climate change, gun control, and more \cite{darwish_unsupervised_2020,stefanov_predicting_2020}. When media narratives are shaped by ideological or commercial agendas, they can distort facts, mislead audiences, and reinforce harmful stereotypes \cite{entman_framing_1993}. Media bias can manifest in various ways, such as through omission, excessive coverage of certain topics, selective presentation of facts, or the use of propaganda strategies that exploit emotions, fears, and prejudices \cite{da_san_martino_fine-grained_2019}. Numerous studies have shown that biased reporting contributes to polarization by presenting information that aligns with audience predispositions while omitting or downplaying counter-evidence \cite{stroud_polarization_2010}. In particular, racial and ethnic minorities are frequently subjected to negative stereotyping in news coverage, with media often over-representing Black and Latino individuals in crime reports and under representing them in positive contexts \cite{dixon_crime_2008, gilens_race_1996}. This distortion fuels public fears, justifies discriminatory policies, and perpetuates systemic racism \cite{dixon_good_2017, romer_treatment_1998, peffley_racial_1997}. Research has also shown that such representations can shape implicit biases and social judgments, contributing to harsher public attitudes toward minority communities \cite{gilliam_prime_2000}. These methods subtly but profoundly influence public opinion, reinforce harmful stereotypes, and perpetuate misinformation, often amplifying social divides and misunderstanding among the public \cite{kumar_decoding_2024}. 

Beyond issues of race, media bias can influence political behavior, erode institutional trust, and affect policy preferences \cite{leeper_political_2014}. Misinformation and selective framing have been linked to declining public trust in journalism, particularly when news appears to align with partisan interests \cite{lazer_science_2018, nyhan_when_2010}. These effects are amplified in digital ecosystems, where algorithmically driven news feeds can further entrench echo chambers and filter bubbles \cite{bakshy_exposure_2015, pariser_filter_2011}. This rapid proliferation not only makes real-time content moderation a difficult task, but also heightens the risk that biased or misleading information may go unchecked and spread rapidly \cite{budak_fair_2016}. Consequently, unchecked media bias not only impairs the public’s ability to make informed decisions but also weakens the media’s role as a democratic watchdog. Identifying media bias has become increasingly important given the widespread dissemination of misinformation and disinformation on social media platforms, which significantly influences public perception and decision-making \cite{yu_classifying_2008, liu_politics_2022}.

Traditional approaches to identifying and mitigating media bias have typically depended heavily on human moderation (e.g., community notes on Twitter/X) and editorial oversight \cite{budak_fair_2016}. While humans can bring critical real-time contextual insight, these approaches inherently introduce subjective biases, inconsistencies, and scalability challenges~\cite{elejalde_nature_2018}. Human moderators' decisions can be influenced by their personal beliefs, experiences, and cultural contexts, leading to uneven enforcement and difficulty in maintaining standardized criteria across large datasets. Furthermore, editorial policies, designed to curb biases, vary significantly between organizations and are often inconsistently applied due to individual interpretations and operational constraints \cite{trhlik_quantifying_2024, budak_fair_2016}.

In recent years, advancements in Natural Language Processing (NLP) have paved the way for greater scalability and more consistent application across extensive digital content \cite{dale_law_2019, velupillai_using_2018}. Notably, transformer-based Large Language Models (LLMs) have demonstrated potential in enhancing the detection of subtle linguistic cues and complex contextual biases that traditional manual and rule-based methods frequently overlook \cite{raza_unlocking_2024, raza_dbias_2024}. Researchers have increasingly relied on powerful large language models (LLMs) as potential tools for predicting media bias \cite{lin_inditag_2024,lin_investigating_2024}. Yet, prior research often treats bias detection and bias mitigation as separate tasks, not only in methodology but also in how their effectiveness is evaluated—frequently using different standards for each \cite{urman_silence_2025,esiobu_robbie_2023}. In contrast, our approach ensures consistency by applying human evaluation to both processes and by using the same bias detection model to assess the outputs of our mitigation system. This framework provides a coherent evaluation, allowing us to examine the limitations of mitigation in the context of the system used to detect bias. Moreover, biases often shift and evolve in response to societal events, political climates, and public discourse; thus, mitigation strategies must also be dynamically adaptable \cite{pansanella_mass_2023}. The literature also suggests without the practical implementation of these computational tools into existing editorial workflows, their practical utility in real-world scenarios is limited and their potential to contribute effectively to media transparency and accountability is diminished \cite{franks_using_2022}.

Addressing these gaps, our research introduces a novel framework that integrates both bias detection and bias mitigation within crime-related news reporting contexts. Our proposed framework aims not only to detect biases with high accuracy but also to systematically reduce them thereby  enhancing the practical applicability and effectiveness of current bias mitigation frameworks common online and in media. Specifically, our contributions are as follows:

\begin{itemize}
\item We conduct a comparative assessment of six LLMs, evaluated through human validation, to determine their ability to detect biased language in crime-related articles.
\item We collect and compile a dataset of over 30,000 crime-related articles, covering a time-frame from 2013 to 2023, providing a foundation for bias analysis. Using this dataset, we conduct a large-scale investigation into biased news coverage drawn from articles published by five news agencies which vary across the United States political spectrum.
\item We investigate the performance of LLMs in rephrasing biased language while maintaining the contextual and narrative coherence of news content and validate their performance with human annotators.
\end{itemize}

By introducing a systematic and scalable AI-driven solution, our research aims to improve bias detection and mitigation in news media, thereby fostering a more balanced, transparent, and informed public discourse environment.

\section{Related Work}
Bias detection in news media has historically relied on manual annotation, with journalists, fact-checkers, and watchdog organizations meticulously assessing articles for biased language, selective framing, or omissions of key information. Studies have shown these manual methods to be adept at identifying nuanced biases, particularly those relying heavily on context and inference \cite{spinde_enhancing_2024}. However, manual approaches inherently suffer from subjectivity, inconsistency across annotators, and significant scalability constraints due to the vast volume of digital news content \cite{trhlik_quantifying_2024, budak_fair_2016}.

To overcome limitations inherent in traditional manual methods, crowd sourced content analyses have emerged, providing scalability improvements through distributed annotation tasks. Despite their benefits, these approaches suffer from significant variability in bias perception across different annotators, reducing consistency and reliability in bias identification \cite{budak_fair_2016}. To address these challenges, automated computational approaches using NLP techniques, including sentiment analysis, entity recognition, and syntactic parsing, have gained prominence due to their scalability and ability to manage extensive datasets efficiently \cite{mozafari_hate_2020}.

Recent research has explored various approaches to mitigating bias through Large Language Models (LLMs), focusing on both static and contextualized embeddings. For static embeddings, methods such as debiasing through projection \cite{bolukbasi_man_2016,ravfogel_null_2020} and gender-neutral embedding learning \cite{zhao_learning_2018} were introduced, though many rely heavily on predefined word lists or external resources \cite{gonen_lipstick_2019}. Kaneko et al. ~\cite{kaneko_dictionary-based_2021} proposed dictionary-based debiasing to overcome this limitation, but its applicability is constrained by dictionary coverage and linguistic variability. 

Bias mitigation strategies fall into categories including pre-processing and in-training methods \cite{gallegos_bias_2024}. Pre-processing focuses on modifying model inputs—such as data and prompts—to enhance representation and reduce bias. This can include techniques such as data augmentation \cite{qian_perturbation_2022}, filtering out biased samples \cite{garimella_demographic-aware_2022}, adjusting prompts \cite{venkit_nationality_2023}, or refining pre-trained representations to be less biased.

In-training on large text corpora is extensively leveraged by studies in transformer-based LLMs, enabling contextual understanding and recognition of subtle biases often missed by simpler NLP methods \cite{kumar_decoding_2024, spinde_exploiting_2022}. For instance, transformer-based multitask learning has demonstrated improvements in detecting nuanced linguistic biases through joint training on related tasks, achieving superior results compared to single-task models \cite{spinde_exploiting_2022}. In contetualized embeddings, efforts to reduce toxicity and social biases include dataset curation \cite{nangia_crows-pairs_2020,nadeem_stereoset_2020}, generative discrimination \cite{krause_gedi_2020}, and debiasing representations \cite{liang_towards_2020}. Nevertheless, this method often require extensive retraining, large datasets, or manually curated interventions \cite{dathathri_plug_2020,gururangan_dont_2020}, limiting their practicality. 

Moreover, attempts to evaluate and mitigate bias at the sentence level \cite{may_measuring_2019,basta_evaluating_2019,kurita_measuring_2019} have yet to produce reliable post-hoc solutions that eliminate bias without retraining, as highlighted by the shortcomings in models proposed by \cite{zhao_gender_2019} and \cite{park_reducing_2018}. Despite their strengths, these advanced models face challenges in consistently aligning their outputs with human judgments of bias and avoiding the introduction of new algorithmic biases. Ongoing research is actively exploring approaches to ensure these advanced models accurately reflect human bias perceptions while minimizing algorithm-induced biases \cite{raza_unlocking_2024}. Overall, despite promising techniques, most existing approaches fall short in real-world applicability due to their reliance on expensive retraining, static interventions, or lack of generalizability.

\section{Approach}

\subsection{Dataset}
Here, we examine media bias by compiling a dataset consisting of over 30,000 crime-related articles published between 2013 and 2023. The dataset was carefully curated from five news publishers, selected to represent a comprehensive political spectrum ranging from liberal to conservative viewpoints. Namely, these publishers include The Daily Beast, CNN, Newsweek, The Washington Times, and Fox News. This selection was informed by established categorizations found in prior media bias research ~\cite{amy_watson_most_2024,gilroy_research_2024}, thus ensuring that each publisher fell under a different segment of the political spectrum. Data acquisition leveraged the ~\cite{wayback_machine_wayback_nodate}, a digital archive that enabled access to historical records of published news articles, allowing us to examine reporting patterns, editorial biases, and linguistic variations associated with crime reporting which may have evolved over the course of more than a decade. By explicitly targeting crime-related journalism, the dataset also allows for a large-scale investigation into racial bias and its potential implications on public perception and attitudes. Each of the 30,000 articles is parsed to extract several components, namely, its publication date, set of author(s), title, and the complete main text. The processed articles were subsequently stored using a JSON-based schema and categorized according to their respective publishers.

\subsection{Bias Detection Methodology}
To detect bias language in news articles, we employed six LLMs, namely: GPT-4o, GPT-4o Mini, Gemini Pro, Gemini Flash, Llama 8B, and Llama 3B. 

First, each article was broken down into its individual paragraphs, amounting to 552,883 paragraphs in total. Each paragraph within the dataset was then assessed by the LLMs, which assigned scores on a three-tier scale: '0' indicating negligible or no bias, '1' signifying moderate bias, and '2' representing extreme bias. To guide these assessments, the LLMs were prompted to identify a number of signals of biased language. These signals included loaded language, selective framing of narratives, emotional appeals, and deliberate or inadvertent omission of critical information. See the Bias detection prompt section in the Appendix for the exact prompt used.

\subsection{Debiasing Framework}
A structured debiasing framework was developed and implemented, involving the following stages:

\begin{itemize}
\item \textbf{Identification}: Utilizing the results from the bias detection phase, paragraphs flagged as biased (scoring either '1' or '2') were systematically identified and cataloged for subsequent processing.
\item \textbf{Mitigation}: GPT-4o Mini was identified as the optimal model for the refinement process, selected based on its performance metrics in the bias detection stage of recognizing and locating bias within paragraphs.
\end{itemize}

The refinement process involved the rephrasing and restructuring of the flagged paragraphs, aimed explicitly at mitigating detected biases while preserving the original informational and contextual integrity. Three different prompts were used to complete the task on three different levels of bias mitigation. Specifically, in addition to mitigating explicit endogenous textual biases (i.e., biased language used by the author of the article) which is considered in the first prompt, the second and third prompts also considered exogenous biases common in journalistic content, such as those found in quotations, citations, and paraphrased segments external to the article. The third prompt, in particular, further emphasizes the usage of neutral and abstract language in relation to emotionally charged phrases. Importantly, such biased language is not necessarily due to the choice of language made by the authors or publisher of the article, but rather the content they chose to cover. Nonetheless, for the purposes of our study, we treat such exogenous stimuli as biased language which is to be identified and addressed by the LLM. Therefore, to ensure the journalistic integrity of the outputted ``debiased'' paragraph, the LLMs were instructed to not directly modify quotes with biased language, and instead, instructed to alter the phrasing of the paragraph as to remove the quoted remarks entirely by paraphrasing it without the biased language used. To be clear, this decision to modify both endogenous and exogenous biases is made under the assumption that debiasing tools, such as the one proposed in this study, are to be used by individuals sensitive to biased language as a whole, rather than biased language stemming from choices made by the authors of a given article. See the Debiasing prompts section in the Appendix for the exact prompts used.

\section{Results}

\subsection{Bias Detection Evaluation}
Our setup was used to evaluate the efficacy of the bias detection models. Each selected article was processed through six LLMs tasked with independently analyzing every paragraph within these articles. The models also provided a justification for their bias assessments, explicitly highlighting the sentences or phrases that contribute to their assigned bias scores. This transparency in scoring was designed to enable human evaluators to understand the rationale behind each model’s judgment.

Subsequently, each scored paragraph was subjected to human evaluation, involving five independent annotators per paragraph. Annotators were recruited through~\cite{prolific_prolific_nodate}. Annotators evaluated the validity of the LLM-generated bias scores and provided their own bias score for each paragraph. This human evaluation process was structured to serve as the ground truth, allowing for accurate benchmarking of each LLM’s performance in bias detection.

Table~\ref{table:llm_detection_performance_table} below details the performance of each the models tested. As shown, GPT-4o Mini demonstrated the best performance overall, with an exact match to the human-majority score 92.5\% of the time.

\begin{table}[htbp!]
\centering
\small
\begin{tabular}{lcccc}
Model                 & Exact Match (\%) & Krippendorff's Alpha & Cohen's Kappa & F2 Score \\ \hline
Gemini 1.5 Flash      & 92.325           & 0.701                & 0.702         & 0.923    \\
Gemini 1.5 Pro        & 90.385           & 0.627                & 0.629         & 0.904    \\
GPT 4o                & 92.375           & 0.641                & 0.641         & 0.924    \\
GPT 4o Mini           & \textbf{92.499}           & \textbf{0.719}                & \textbf{0.721}         & \textbf{0.925}    \\
Llama 3.2 3B Instruct & 85.993           & 0.527                & 0.535         & 0.860    \\
Llama 3.1 8B Instruct & \textbf{92.499}           & 0.664                & 0.664         & \textbf{0.925}   \\ \hline
\end{tabular}
\caption{Performance metrics of the six different LLMs tested for the purpose of bias detection. Each metric is computed against the majority vote of human-annotators.}
\label{table:llm_detection_performance_table}
\end{table}

\subsection{Temporal, geographical, and publisher variations in biased media coverage}

\begin{figure}[htbp!]
    \centering
    \includegraphics[width=1\linewidth]{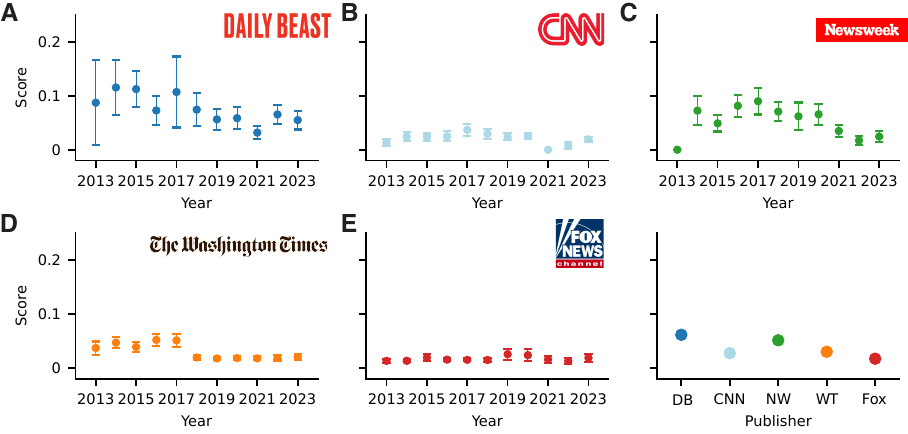}
    \caption{(\textbf{A - E}) The average bias score of articles from each publisher over time. (\textbf{F}) The average bias score for each publisher overall.}
    \label{fig:bias_by_publisher}
\end{figure}

Next, we look to understand the prevalence of biased language in news media over time. We begin by examining the average bias score per paragraph for each publisher across the decade of news coverage available in our dataset, the results of which are illustrated in Figure~\ref{fig:bias_by_publisher}. As can be seen, there were no significant changes in temporal trends with regards to the propensity for biased coverage across the different publishers (see Figures~\ref{fig:bias_by_publisher}A - E). Comparing publishers overall, we find statistically significant differences across publishers. Namely, we find that the DailyBeast were most likely to publish paragraphs with biased language (Independent t-test; $p < 0.001$ for all pairwise comparisons), followed by Newsweek (Independent t-test; $p < 0.001$ for all non DailyBeast pairwise comparisons).

\begin{figure}[t!]
    \centering
    \includegraphics[width=\linewidth]{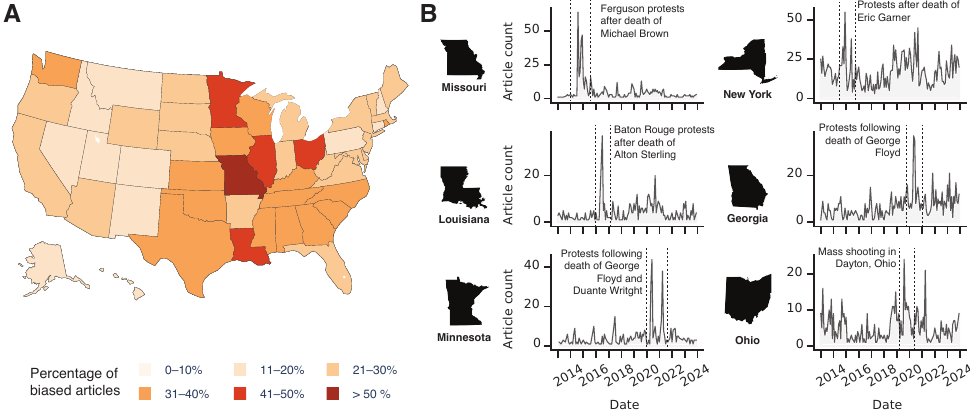}
    \caption{\textbf{Biased coverage of crime in the United States.} (\textbf{A}) A heatmap illustrating the proportion of articles covering a crime in a given state with biased language. (\textbf{B}) For the states of Missouri, Louisiana, Minnesota, New York, Georgia, and Ohio, the number of articles with biased language in each year, with information detailing relevant social issues corresponding to spikes in biased coverage.}
    \label{fig:figure_1}
\end{figure}

When comparing bias scores geographically, several trends emerge, with certain states being covered less but exhibiting systematically higher bias levels when covered (see Table~\ref{table:state_article_counts} for the number of articles covering crimes in a given state).  While states with larger populations—such as California, Texas, and Florida—tend to illicit more articles overall, they do not consistently exhibit strong correlations between real-world events and increases in media bias. In contrast, states with lower media coverage overall, including Missouri, Louisiana, and Minnesota, show more dramatic spikes in biased reporting as they tend to be specifically covered during periods of civil unrest or violence. Figure~\ref{fig:figure_1}A illustrates the proportion of articles with biased language covering a crime which occurred in a given state. Although these states are mentioned less frequently in national news, the share of biased articles among their coverage rises significantly during certain high-profile events. Investigating this trend temporally, we find that these elevated bias rates are largely attributed to significant spikes in paragraphs containing biased language surrounding major protests and events with civil unrest. As can be seen in Figure~\ref{fig:figure_1}B, biased coverage in such states centered around major instances of civil unrest. For instance, we see a major spike in the state of Missouri during 2014, which corresponded to articles with biased language surrounding the Ferguson protests after the death of Michael Brown~\cite{the_new_york_times_missouri_2014} . We see similar instances in the states of Louisiana (Baton Rouge protests in 2016~\cite{abc_news_protests_2016}), Minnesota (George Floyd protests in 2020~\cite{bbc_news_george_2020} and Daunte Wright protests in 2021~\cite{the_new_york_times_what_2022}), as well as other instances in New York~\cite{the_guardian_eric_2014}, Georgia~\cite{cnbc_protests_2020}, and Ohio~\cite{cbs_news_ohio_2019}. Taken together, these patterns affirm that biased media coverage is not evenly distributed across states or over time, but are rather instead tightly linked to specific socio-political incidents. These results underscore the utility of LLM-based bias detection systems for capturing fine-grained, event-driven shifts in media discourse—offering a scalable and interpretable tool for researchers, journalists, and policy analysts concerned with media transparency and accountability.

\begin{figure}[htb!]
    \centering
    \includegraphics[width=\linewidth]{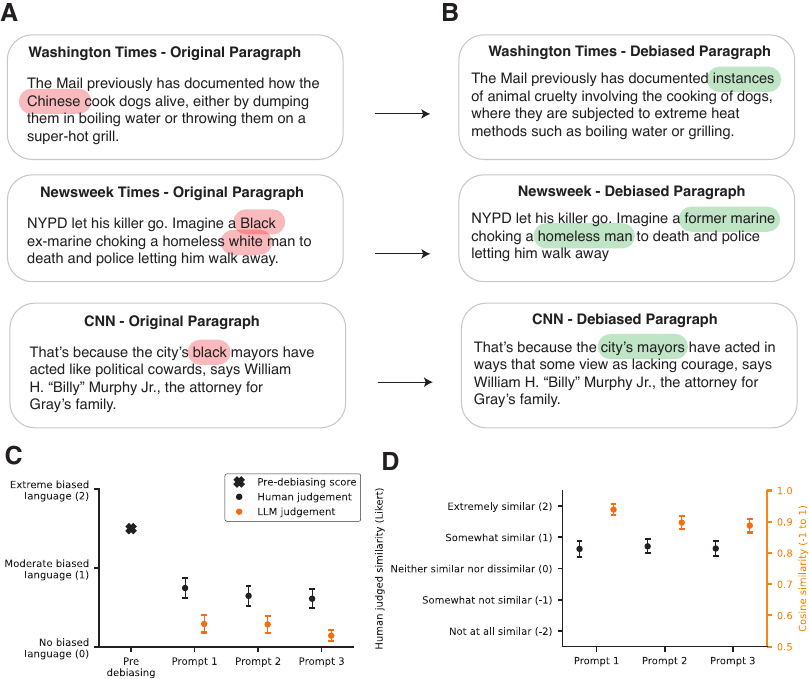}
    \caption{\textbf{Debiasing effectiveness} (\textbf{A, B}) An illustrative example of a paragraph which includes biased language (A), and after the paragraph is modified to remove biased language (B). (\textbf{C}) The average bias score of articles before and after debiasing as judged by human annotators (black circles) and by GPT-4o-mini (orange circles). (\textbf{D}) The similarity of biased and debiased paragraphs as determined by humans (left y-axis, black circles), and their cosine similarity scores (right y-axis, orange circles).}
    \label{fig:fig2}
\end{figure}

\subsection{Debiasing Evaluation}

Based on the aforementioned performance evaluation, GPT-4o Mini emerged as the best-performing model for bias detection (see Table~\ref{table:llm_detection_performance_table}) and was thus selected for the debiasing process. Using three debiasing prompts, biased paragraphs (those previously scoring '1' or '2') were reprocessed. These debiasing prompts specifically guided the LLM towards language neutrality and bias minimization to three different levels of bias mitigation, while maintaining the factual accuracy and original contextual information of the paragraphs. Figure~\ref{fig:fig2}A and ~\ref{fig:fig2}B present illustrative examples of original and debiased paragraphs, respectively.

Post-debiasing, the refined paragraphs were reassessed through two validation steps, involving both LLM and human evaluators. 
Specifically, each of the debiased paragraphs were evaluated by five independent human annotators (also recruited via Prolific). Annotators were asked two complete two sets of tasks. The first task mirrored the first human annotation task detailed in the previous section, where annotators were asked to determine the level of bias included in the paragraph. In parallel, GPT-4o Mini was prompted with the same question when given the debiased paragraph. The results of this analysis can be seen in Figure~\ref{fig:fig2}C. As can be seen, each of the three prompts significantly reduced the bias contained in the articles, as judged by humans (Independent t-test; $p < 0.001$ for all prompts) and by GPT-4o Mini (Independent t-test; $p < 0.001$ for all prompts).

\begin{table}[htbp!]
\centering
\footnotesize
\begin{tabular}{lcccc}
         & \multicolumn{2}{c}{Human Judgement}                                                                                                                                                 & \multicolumn{2}{c}{LLM Judgement}                                                                                                                                                                                           \\ \hline
         & \begin{tabular}[c]{@{}c@{}}\% of moderate bias\\ articles with no\\ remaining bias\end{tabular} & \begin{tabular}[c]{@{}c@{}}\% of extreme bias\\ articles with no\\ remaining bias\end{tabular} & \multicolumn{1}{c}{\begin{tabular}[c]{@{}c@{}}\% of moderate bias\\ articles with no\\ remaining bias\end{tabular}} & \multicolumn{1}{c}{\begin{tabular}[c]{@{}c@{}}\% of extreme bias\\ articles with no\\ remaining bias\end{tabular}} \\ \hline
Prompt 1 & 55.2                                                                                     & 32.1                                                                                     &   84.4                                                                                                &            74.4                                                                                                  \\
Prompt 2 & 55.4                                                                                     & 44.2                                                                                     &                84.8                                                                                              &   72.0                                                                                                           \\
Prompt 3 & 56.4                                                                                     & 44.8                                                                                     &  90.0                                                                                                            &         86.4            \\ \hline                                                                                        
\end{tabular}
\caption{The proportion of moderate bias and extreme bias articles which successfully became unbiased after the debiasing process based on human judgement and LLM judgement.}
\label{table:debiasing_performance_table}
\end{table}

Table~\ref{table:debiasing_performance_table} lists the proportion of paragraphs which were deemed to include no biased language after the debiasing process. As can be seen, of the paragraphs which originally were deemed to include extreme biased language, the best performing prompt successfully eliminated all biased language in 44.8\% of paragraphs. 

The second task involved presenting both the original paragraph as well as the debiased paragraph to the annotator, where the annotator would judge the level of similarity between the two paragraphs. In parallel, we also utilized the ``\textit{all-mpnet-base-v2}''~\cite{hugging_face_sentence-transformersall-mpnet-base-v2_2024} embedding model to extract vector embeddings of the original-debiased paragraph pairs. ``\textit{all-mpnet-base-v2}'' inherits the advantages of the traditional embedding models while accounting for auxiliary position information as input, achieving better results
on these tasks compared with previous state-
of-the-art pre-trained methods (e.g., BERT,
XLNet, RoBERTa)~\cite{song2020mpnetmaskedpermutedpretraining}. From there, we compute the cosine similarity between each pair of paragraphs. The results of these analyses are presented in Figure~\ref{fig:fig2}D, with specific values listed in  Table~\ref{table:debiasing_similarity_table}. Here, we see that, according to human judgement, the second prompt produced paragraphs which best maintained the contextual similarity of the respective original paragraphs. In contrast, with regards to cosine similarity, the first prompt, which required the fewest text modifications, naturally yielded the highest cosine similarity.

\begin{table}[htbp!]
\centering
\begin{tabular}{lcc}
         & \begin{tabular}[c]{@{}c@{}}Average Similarity Score\\ (Scale -2 to 2)\end{tabular} & \begin{tabular}[c]{@{}c@{}}Cosine Similarity Score\\ (Scale -1 to 1)\end{tabular} \\ \hline
Prompt 1 & 0.629                                                                              & 0.871                                                                             \\
Prompt 2 & 0.712                                                                              & 0.837                                                                             \\
Prompt 3 & 0.643                                                                              & 0.797                                   \\ \hline                                         
\end{tabular}
\caption{The average contextual similarity score (based on human annotators), and cosine similarity score, between biased and debiased paragraphs for each prompt.}
\label{table:debiasing_similarity_table}
\end{table}

\section{Discussion and Conclusion}


In this study, we evaluated the ability of LLMs to detect, and subsequently mitigate, biased language in news articles, focusing on crime-related reporting from five politically diverse news sources over the course of a decade. Our results indicate that LLMs do exhibit strong performance in detecting bias, with the best performing model, GPT-4o Mini, having an accuracy of 92.5\% when compared to human annotators. Using this bias detection mechanism, we studied geographical and temporal variations in media bias, and found that biases tended to correlate with socio-political events. Specifically, states with lower media coverage, such as Missouri, Louisiana, and Minnesota, showed increased biased reporting during periods of civil unrest. Lastly, we show that an LLM-driven bias mitigation process is effective in reducing biased language while simultaneously maintaining the relevant contextual information present in news articles. 

Our study is not without limitations. First, the dataset, while extensive, is limited to crime-related articles from five news sources, which may not fully represent broader news reporting. In addition, our study focuses on crime-related articles in the United States specifically. As such, future research can expand the dataset to include both more diverse topics, news sources, and international contexts. Second, the usage of LLMs for bias detection and debiasing introduces potential algorithmic biases, as these models are trained on large datasets that may perpetuate existing societal biases. There is no shortage of existing research highlighting such concerns, in both the context of image~\cite{aldahoul2024ai, lopez2025generative, fraser2024examining} and text generation~\cite{salinas2024s, an2024measuring, an2025measuring}. Furthermore, identifying bias is fundamentally a subjective endeavor, even at the human-level.

The aforementioned limitations offer several promising directions for future research. First, while our study looks to examine bias broadly, and identifies specific socio-cultural events which correlate with spikes in biased coverage, future work may examine the specific parties (e.g., specific racial, gender, or political groups) which are most susceptible to biased language. Second, to address the concern of algorithmic biases, scaling the framework through the incorporation of adversarial training and hybrid methodologies may improve framework robustness and fairness. Combining deep learning approaches with explicit linguistic or rule-based systems could further enhance interpretability and reduce reliance on black-box model outputs.

Lastly, translating this framework into real-world applications represents the most impactful path forward. On the production side, AI-assisted authoring tools could be developed to provide journalists with real-time feedback on potentially biased language during the writing process, enabling more reflective editorial practices. On the consumption side, extending this framework to user-facing tools—such as browser extensions—could allow those particularly sensitive to biased language to have agency in the type of media they consume. Moreover, it could allow readers to engage more critically with media content by dynamically flagging biased language and offering alternative framings. Incorporating user feedback into these tools would further support iterative refinement, improving both detection accuracy and mitigation effectiveness over time. Pursuing these directions would contribute not only to the academic development of bias-aware language technologies, but also to their practical deployment in fostering fairness, transparency, and critical engagement within the broader media ecosystem.

\section*{Ethics Statement}
The research was approved by the authors' Institutional Review Board (HRPP-2025-10). All research was performed in accordance with relevant guidelines and regulations. Informed consent was obtained from all participants in every segment of this study.

\bibliographystyle{naturemag}
\bibliography{colm2025_conference}

\appendix
\section{Appendix}

\subsection{Prompts}

\subsubsection{Bias detection prompt}

\begin{quote}
    Instructions:

    1. Topic Identification: Identify the main topic of the text in one or two words.
    2. Bias Detection: Assess whether the writing styles or the narrative exhibit any form of bias against certain racial or religious groups (e.g. African Americans, LatinX or Hispanics, Asians, Muslims, immigrants, native Americans or indigenous people, refugees, Middle Eastern, minorities, and other marginalized racial groups you can think of). Choose either 'yes' or 'no'. Do not choose anything else.
    3. Bias Scoring: Rate the level of bias on a scale of 0-2.
        \begin{itemize}
            \item Level 0: Negligible bias or no bias detected. Content shows balanced representation and fair treatment of all racial groups, using neutral language and avoiding stereotypes.
            \item Level 1: Moderate bias. Content contains subtle stereotypes, generalizations about racial groups, or imbalanced representation, but without overtly discriminatory language.
            \item Level 2: Extreme bias. Content exhibits explicit racial prejudice, uses derogatory language, promotes harmful stereotypes, or contains clear discriminatory statements about racial groups.
        \end{itemize}
    4. Evidence Identification: Extract specific full sentences that exhibit bias, separating them by '\&'. Avoid selecting partial or paraphrased sentences.
    5. Bias Justification: Provide a brief explanation for why each sentence is considered biased, including references to stereotypical terms, exclusionary language, or specific tone, separated by '\&'.
    6. Bias Summary: Summarize the overall reason for the bias in less than 5 words.
    7. Bias Removal: Briefly explain how you rewrite the biased sentence to mitigate the bias.

    Note that only detect race related bias in crime. IGNORE other biases that are not related to race and report no bias detected.

    Output the result as a valid JSON object with the following fields:

    \begin{verbatim}
    {
        "Main Topic": "[chosen topic]",
        "Bias Detected": "[yes/no]",
        "Bias Score": "[0-2]",
        "Biased Sentences": "[sentences separated by '&']",
        "Bias Justification": "[justifications separated by '&']",
        "Bias Summary": "[summary in less than 5 words]",
        "Bias Removal": "[how you would mitigate biases]"
    }
    \end{verbatim}
\end{quote}

\subsubsection{Debiasing prompts}

\textbf{Prompt 1}:

\begin{quote}
    Instructions:

    1. Input Analysis: Carefully analyze the full paragraph given that contains racial bias against certain racial or religious groups (e.g. African Americans, LatinX or Hispanics, Asians, Muslims, immigrants, native Americans or indigenous people, refugees, Middle Eastern, minorities, and other marginalized racial groups you can think of) in crime reporting.

    2. Preservation Requirements: Identify factual information that must be preserved in the Rewritten Full Paragraph:
        \begin{itemize}
            \item Core events and actions
            \item Relevant context and details
            \item Verified facts and statistics
        \end{itemize}

    3. Output Generation: Provide the complete Rewritten Full Paragraph that:
        \begin{itemize}
            \item Maintains all factual information from the original
            \item Removes identified racial biases
            \item Apply neutral descriptors that avoid racial connotations
            \item Remove emotional language that reinforces racial stereotypes
            \item Preserves the original meaning and intent where appropriate
            \item Reads naturally and maintains coherent flow
        \end{itemize}

    Output the result as a valid JSON object with the following fields:

    \begin{verbatim}
    {
        "Rewritten Full Paragraph": "[remove bias for
        the paragraph only]",
        "Transformation Summary": "[explanation of changes
        made in less than 5 words]",
        "Preservation Analysis": "[explanation of how factual
        information was preserved]"
    }
    \end{verbatim}
\end{quote}

\textbf{Prompt 2}:

\begin{quote}
    Instructions:

    1. Input Analysis: Carefully analyze the full paragraph given that contains racial bias against certain racial or religious groups (e.g. African Americans, LatinX or Hispanics, Asians, Muslims, immigrants, native Americans or indigenous people, refugees, Middle Eastern, minorities, and other marginalized racial groups you can think of) in crime reporting.

    2. Preservation Requirements: Identify factual information that must be preserved in the rewritten Full Paragraph:
        \begin{itemize}
            \item Core events and actions
            \item Relevant context and details
            \item Verified facts and statistics
        \end{itemize}

    3. Quote and Citied Material: When encountering quotes or other cited material that contains bias:
        \begin{itemize}
            \item Use indirect attribution to convey the content without the biased language (e.g., "The witness described the suspects..." instead of directly quoting biased descriptions)
            \item Apply selective quoting by only using direct quotes for unbiased portions and paraphrasing biased segments
            \item For article titles with inflammatory or biased language, describe the general topic of the article instead of quoting the full title (e.g., "He shared an article with anti-Muslim content" instead of reproducing a biased headline)
        \end{itemize}

    4. Output Generation: Provide the complete Rewritten Full Paragraph that:
        \begin{itemize}
            \item Maintains all factual information from the original
            \item Removes identified racial biases
            \item Apply neutral descriptors that avoid racial connotations
            \item Remove emotional language that reinforces racial stereotypes
            \item Preserves the original meaning and intent where appropriate
            \item Reads naturally and maintains coherent flow
        \end{itemize}

    Output the result as a valid JSON object with the following fields:

    \begin{verbatim}
    {
        "Rewritten Full Paragraph": "[remove bias for the
        paragraph only]",
        "Transformation Summary": "[explanation of changes
        made in less than 5 words]",
        "Preservation Analysis": "[explanation of how
        factual information was preserved]"
    }
    \end{verbatim}
\end{quote}

\textbf{Prompt 3}:

\begin{quote}
    Instructions:

    1. Input Analysis: Carefully analyze the full paragraph given that contains racial bias against certain racial or religious groups (e.g. African Americans, LatinX or Hispanics, Asians, Muslims, immigrants, native Americans or indigenous people, refugees, Middle Eastern, minorities, and other marginalized racial groups you can think of) in crime reporting.

    2. Preservation Requirements: Identify factual information that must be preserved in the Rewritten Full Paragraph:
        \begin{itemize}
            \item Core events and actions
            \item Relevant context and details
            \item Verified facts and statistics
        \end{itemize}

    3. Quote and Title Handling: When encountering quotes, article titles, or other cited material that contains bias:
        \begin{itemize}
            \item Completely reformulate biased article titles without directly quoting them (e.g., "He shared an article containing inflammatory content" instead of reproducing a biased headline)
            \item Use indirect attribution and focus on behavior rather than identity or target group (e.g., "The individual made inappropriate comments" instead of "The individual used racial slurs toward [group]")
            \item Apply selective quoting by only using direct quotes for unbiased portions and paraphrasing biased segments
            \item Avoid repeating or closely paraphrasing charged terminology even when describing it
        \end{itemize}

    4. Language Selection:
        \begin{itemize}
            \item Use neutral, factual language that avoids both explicit and implicit references to race, ethnicity, or religion when describing negative actions
            \item Focus on actions and behaviors rather than motivations when those motivations involve bias
            \item Abstract references to highly charged incidents, movements, or figures when they carry strong racial connotations
        \end{itemize}

    5. Output Generation: Provide the complete Rewritten Full Paragraph that:
        \begin{itemize}
            \item Maintains all factual information from the original
            \item Removes identified racial biases
            \item Apply neutral descriptors that avoid racial connotations
            \item Remove emotional language that reinforces racial stereotypes
            \item Preserves the original meaning and intent where appropriate
            \item Reads naturally and maintains coherent flow
        \end{itemize}

    Output the result as a valid JSON object with the following fields:

    \begin{verbatim}
    {
        "Rewritten Full Paragraph": "[remove bias for
        the paragraph only]",
        "Transformation Summary": "[explanation of changes
        made in less than 5 words]",
        "Preservation Analysis": "[explanation of how
        factual information was preserved]",
        "Contain Cited Materials":"[does the original
        paragraph contains quotes or cited materials?]: yes/no"
    }
    \end{verbatim}
\end{quote}

\subsection{Supplementary Tables}

\begin{table}[htbp!]
\centering
\footnotesize
\begin{tabular}{lccccccccccc}
Publisher        & 2013 & 2014 & 2015 & 2016 & 2017 & 2018 & 2019 & 2020 & 2021 & 2022 & 2023 \\ \hline
CNN              & 1.41 & 2.57 & 2.19 & 2.38 & 2.84 & 2.41 & 2.3  & 2.56 & 0.0  & 0.71 & 1.83 \\
DailyBeast       & 5.82 & 7.16 & 7.46 & 5.35 & 5.1  & 4.06 & 4.07 & 5.09 & 2.73 & 4.8  & 4.47 \\
Fox News         & 1.13 & 1.48 & 1.8  & 1.41 & 1.49 & 1.28 & 2.59 & 2.56 & 1.49 & 1.24 & 1.63 \\
Newsweek         & 0.0  & 8.35 & 4.41 & 7.14 & 6.61 & 5.21 & 3.81 & 4.96 & 2.74 & 1.32 & 2.13 \\
Washington Times & 2.2  & 4.06 & 3.62 & 4.12 & 4.13 & 2.19 & 2.07 & 2.16 & 2.07 & 1.53 & 1.93 \\ \hline
Overall          & 1.54 & 2.94 & 3.44 & 3.03 & 2.91 & 2.51 & 2.52 & 2.85 & 2.22 & 2.07 & 2.08 \\\hline
\end{tabular}
\caption{The proportion of articles by a given publisher which contain biased language in each year.}
\label{table:publisher_bias_yearly}
\end{table}

\begin{table}[htbp!]
\footnotesize
\centering
\begin{tabular}{lllllllllllll}
\hline
state          & 2013   & 2014   & 2015   & 2016   & 2017   & 2018   & 2019   & 2020 & 2021   & 2022   & 2023 & Mean \\ \hline
Alabama        & $\sim$ & 2.82   & 2.85   & 3.11   & 2.09   & 3.58   & 2.48   & 3.65 & 0.92   & 3.46   & 2.64 & 2.76  \\
Alaska         & $\sim$ & 0.39   & 0.65   & 1.55   & 1.74   & 1.60   & 0.93   & 0.74 & 0.47   & 1.52   & 0.33 & 0.99  \\
Arizona        & 1.35   & 2.36   & 1.88   & 1.98   & 2.14   & 2.60   & 2.73   & 1.23 & 1.37   & 1.50   & 2.05 & 1.93  \\
Arkansas       & 0.52   & 0.53   & 1.24   & 3.28   & 1.79   & 4.14   & 4.90   & 3.12 & 2.41   & 0.72   & 0.76 & 2.13  \\
California     & 0.68   & 2.41   & 4.50   & 3.08   & 2.26   & 2.61   & 2.96   & 2.42 & 2.73   & 2.24   & 1.71 & 2.51  \\
Colorado       & 0.93   & 3.75   & 1.98   & 1.46   & 0.67   & 1.37   & 0.72   & 3.06 & 1.34   & 1.28   & 1.32 & 1.63  \\
Connecticut    & 0.22   & 0.64   & 1.54   & 1.64   & 0.33   & 0.72   & 0.84   & 0.58 & 1.05   & 0.76   & 0.80 & 0.83  \\
Delaware       & 1.92   & 1.13   & 3.21   & 0.45   & $\sim$ & 0.34   & 1.03   & 1.85 & 0.90   & 0.67   & 4.81 & 1.63  \\
Florida        & 3.45   & 3.36   & 2.89   & 2.74   & 1.64   & 2.55   & 1.82   & 2.73 & 1.95   & 1.37   & 3.37 & 2.53  \\
Georgia        & 0.24   & 1.10   & 2.97   & 3.71   & 1.76   & 2.31   & 3.73   & 5.80 & 4.77   & 5.10   & 2.41 & 3.08  \\
Hawaii         & 0.93   & 1.98   & 0.42   & 0.60   & 1.17   & 0.19   & 2.66   & 1.71 & 3.55   & 0.40   & 0.57 & 1.29  \\
Idaho          & 0.99   & 3.89   & 1.23   & 2.59   & 1.53   & 5.11   & 2.31   & 0.90 & 3.59   & 0.22   & 0.37 & 2.07  \\
Illinois       & 1.53   & 5.99   & 3.13   & 1.88   & 1.36   & 1.63   & 2.30   & 2.64 & 3.96   & 0.57   & 2.31 & 2.48  \\
Indiana        & 0.81   & 1.38   & 2.00   & 4.44   & 1.77   & 1.85   & 2.10   & 3.45 & 1.61   & 1.12   & 1.40 & 1.99  \\
Iowa           & 0.30   & 1.12   & 4.16   & 1.73   & 1.18   & 8.89   & 4.49   & 3.39 & 2.94   & $\sim$ & 3.84 & 3.20  \\
Kansas         & 2.75   & 3.29   & 2.97   & 2.44   & 4.23   & 1.62   & 2.70   & 1.75 & 2.90   & 3.44   & 7.88 & 3.27  \\
Kentucky       & 2.38   & 1.34   & 3.04   & 2.44   & 1.77   & 1.43   & 2.11   & 3.32 & 1.70   & 1.11   & 0.86 & 1.95  \\
Louisiana      & 0.34   & 3.98   & 1.91   & 5.20   & 4.73   & 2.36   & 2.54   & 2.97 & 1.86   & 3.10   & 1.52 & 2.77  \\
Maine          & $\sim$ & 0.72   & 0.68   & 1.68   & 0.30   & 1.00   & 1.69   & 1.33 & 0.70   & $\sim$ & 1.44 & 1.06  \\
Maryland       & 1.35   & 1.50   & 3.12   & 0.88   & 6.33   & 1.68   & 4.73   & 6.38 & 1.30   & 1.01   & 2.71 & 2.82  \\
Massachusetts  & 1.49   & 1.15   & 2.53   & 1.72   & 1.97   & 1.24   & 1.65   & 2.52 & 1.43   & 1.47   & 1.90 & 1.73  \\
Michigan       & 1.34   & 3.99   & 5.18   & 1.27   & 2.89   & 1.88   & 2.99   & 5.21 & 0.86   & 1.92   & 1.42 & 2.63  \\
Minnesota      & 0.69   & 1.94   & 1.98   & 5.97   & 3.45   & 3.59   & 3.56   & 3.25 & 3.76   & 3.39   & 3.09 & 3.15  \\
Mississippi    & 1.81   & 2.60   & 1.32   & 2.42   & 2.23   & 0.90   & 2.60   & 4.13 & 1.66   & 1.08   & 5.96 & 2.43  \\
Missouri       & 1.58   & 7.70   & 6.45   & 5.66   & 7.12   & 2.66   & 1.62   & 3.36 & 2.74   & 2.17   & 7.21 & 4.39  \\
Montana        & 1.43   & $\sim$ & $\sim$ & 0.30   & $\sim$ & 0.32   & 0.49   & 1.57 & 2.18   & 1.05   & 0.20 & 0.94  \\
Nebraska       & $\sim$ & 0.72   & 2.46   & 0.58   & 1.90   & 5.21   & 5.58   & 3.16 & 0.41   & 3.25   & 9.78 & 3.30  \\
Nevada         & 0.67   & 1.33   & 3.48   & 0.35   & 1.15   & 0.90   & 4.69   & 1.22 & 1.30   & 2.19   & 0.53 & 1.62  \\
New hampshire  & 0.40   & 1.26   & 0.85   & 3.88   & $\sim$ & $\sim$ & 0.64   & 6.02 & 0.92   & $\sim$ & 0.93 & 1.86  \\
New jersey     & 1.30   & 1.74   & 2.12   & 2.48   & 4.23   & 1.32   & 1.98   & 2.22 & 1.50   & 1.94   & 1.39 & 2.02  \\
New mexico     & 0.68   & 1.46   & 0.58   & 0.23   & 0.28   & 1.06   & 0.81   & 1.40 & 1.41   & 2.70   & 0.74 & 1.03  \\
New york       & 2.56   & 5.21   & 3.90   & 4.71   & 4.18   & 1.94   & 2.71   & 3.02 & 2.19   & 1.97   & 2.79 & 3.20  \\
North carolina & 0.44   & 1.42   & 5.79   & 3.94   & 4.82   & 2.51   & 2.64   & 5.20 & 2.07   & 2.65   & 3.05 & 3.14  \\
North dakota   & 2.04   & 0.56   & $\sim$ & 0.62   & 0.80   & 0.52   & $\sim$ & 1.21 & 2.52   & 0.53   & 1.31 & 1.12  \\
Ohio           & 0.80   & 3.78   & 5.91   & 2.63   & 3.10   & 2.47   & 2.22   & 3.19 & 3.96   & 3.26   & 3.56 & 3.17  \\
Oklahoma       & 4.92   & 4.57   & 5.17   & 6.85   & 3.23   & 3.75   & 4.51   & 4.66 & 2.35   & 1.14   & 2.79 & 3.99  \\
Oregon         & 2.64   & 0.68   & 1.82   & 1.47   & 5.78   & 2.75   & 2.22   & 2.52 & 5.82   & 0.47   & 0.41 & 2.42  \\
Pennsylvania   & 0.65   & 0.80   & 2.09   & 2.09   & 0.96   & 1.36   & 3.66   & 1.13 & 2.16   & 1.15   & 1.80 & 1.62  \\
Rhode island   & 0.37   & 3.73   & 2.51   & 0.67   & 0.62   & 1.00   & 0.44   & 1.43 & 4.07   & 0.33   & 1.21 & 1.49  \\
South carolina & 1.92   & 2.26   & 6.58   & 6.35   & 5.86   & 1.81   & 2.59   & 6.58 & 1.55   & 0.43   & 1.82 & 3.43  \\
South dakota   & $\sim$ & $\sim$ & $\sim$ & $\sim$ & 3.08   & $\sim$ & 1.80   & 1.69 & 2.27   & $\sim$ & 1.19 & 2.01  \\
Tennessee      & $\sim$ & 3.66   & 2.39   & 3.40   & 1.47   & 1.54   & 3.08   & 3.12 & 2.60   & 0.79   & 2.48 & 2.45  \\
Texas          & 2.32   & 2.10   & 4.05   & 3.66   & 3.03   & 2.17   & 3.82   & 2.86 & 2.70   & 1.77   & 3.68 & 2.92  \\
Utah           & 0.41   & 3.86   & 1.85   & 0.25   & 0.60   & 1.12   & 0.53   & 1.72 & 0.60   & 0.41   & 0.70 & 1.10  \\
Vermont        & 0.79   & 0.98   & 0.22   & $\sim$ & 0.47   & 1.10   & 4.18   & 1.60 & $\sim$ & 0.29   & 5.20 & 1.65  \\
Virginia       & 1.86   & 0.82   & 3.25   & 1.78   & 4.07   & 3.02   & 1.65   & 1.89 & 2.34   & 0.89   & 0.85 & 2.04  \\
Washington     & 2.15   & 3.45   & 4.69   & 3.11   & 3.99   & 2.44   & 3.97   & 3.38 & 2.26   & 0.92   & 1.53 & 2.90  \\
West virginia  & 0.51   & $\sim$ & 1.19   & 2.11   & 1.90   & 5.98   & 2.36   & 0.65 & 1.08   & $\sim$ & 0.24 & 1.78  \\
Wisconsin      & 2.78   & 6.65   & 2.42   & 4.12   & 3.11   & 2.50   & 1.24   & 3.87 & 3.68   & 1.49   & 2.19 & 3.10  \\
Wyoming        & 0.97   & $\sim$ & $\sim$ & $\sim$ & 0.65   & $\sim$ & 3.25   & 0.94 & 0.48   & 0.78   & 4.25 & 1.62  \\ \hline
Mean        & 1.37   & 2.44   & 2.76   & 2.54   & 2.46   & 2.23   & 2.52   & 2.75 & 2.14   & 1.56   & 2.35 & 2.26  \\ \hline
\end{tabular}
\caption{The proportion of paragraphs about a crime occurring in a given state which contain biased language in each year.}
\label{table:state_bias_yearly}
\end{table}

\begin{table}[htbp!]
\centering
\footnotesize
\begin{tabular}{lcccccccccccc}\hline
state          & 2013 & 2014 & 2015 & 2016 & 2017 & 2018 & 2019 & 2020 & 2021 & 2022 & 2023 & Total \\
Alabama        & 0    & 6    & 5    & 7    & 12   & 15   & 13   & 6    & 7    & 11   & 21   & 103   \\
Alaska         & 0    & 1    & 1    & 5    & 4    & 4    & 5    & 2    & 1    & 4    & 1    & 28    \\
Arizona        & 10   & 9    & 6    & 11   & 19   & 24   & 15   & 15   & 11   & 8    & 18   & 146   \\
Arkansas       & 2    & 2    & 3    & 7    & 4    & 14   & 17   & 10   & 3    & 3    & 3    & 68    \\
California     & 22   & 34   & 59   & 64   & 35   & 74   & 85   & 62   & 58   & 47   & 54   & 594   \\
Colorado       & 17   & 12   & 4    & 5    & 5    & 13   & 8    & 20   & 14   & 12   & 11   & 121   \\
Connecticut    & 3    & 2    & 2    & 5    & 4    & 4    & 6    & 4    & 1    & 3    & 2    & 36    \\
Delaware       & 3    & 1    & 2    & 2    & 0    & 2    & 3    & 6    & 3    & 2    & 2    & 26    \\
Florida        & 42   & 39   & 30   & 39   & 42   & 48   & 37   & 34   & 24   & 15   & 57   & 407   \\
Georgia        & 0    & 8    & 7    & 19   & 12   & 13   & 12   & 58   & 28   & 16   & 13   & 186   \\
Hawaii         & 1    & 3    & 1    & 1    & 3    & 0    & 0    & 0    & 2    & 0    & 0    & 11    \\
Idaho          & 4    & 5    & 2    & 10   & 8    & 7    & 3    & 2    & 3    & 0    & 2    & 46    \\
Illinois       & 4    & 5    & 30   & 12   & 10   & 13   & 14   & 19   & 17   & 3    & 15   & 142   \\
Indiana        & 4    & 6    & 4    & 10   & 19   & 9    & 15   & 15   & 22   & 24   & 19   & 147   \\
Iowa           & 0    & 2    & 8    & 4    & 0    & 12   & 7    & 6    & 10   & 0    & 0    & 49    \\
Kansas         & 5    & 6    & 7    & 9    & 16   & 9    & 4    & 3    & 6    & 6    & 13   & 84    \\
Kentucky       & 1    & 1    & 2    & 3    & 5    & 5    & 2    & 27   & 4    & 4    & 7    & 61    \\
Louisiana      & 0    & 1    & 3    & 45   & 22   & 13   & 11   & 20   & 3    & 9    & 9    & 136   \\
Maine          & 0    & 1    & 0    & 2    & 0    & 2    & 3    & 4    & 1    & 0    & 6    & 19    \\
Maryland       & 3    & 5    & 16   & 6    & 27   & 15   & 17   & 8    & 4    & 4    & 9    & 114   \\
Massachusetts  & 10   & 6    & 17   & 7    & 6    & 7    & 3    & 4    & 3    & 6    & 4    & 73    \\
Michigan       & 3    & 7    & 6    & 3    & 10   & 10   & 3    & 13   & 5    & 8    & 6    & 74    \\
Minnesota      & 0    & 3    & 5    & 18   & 22   & 1    & 9    & 32   & 21   & 10   & 7    & 128   \\
Mississippi    & 1    & 3    & 2    & 3    & 5    & 5    & 5    & 5    & 3    & 4    & 14   & 50    \\
Missouri       & 0    & 120  & 67   & 21   & 29   & 6    & 5    & 10   & 3    & 3    & 1    & 265   \\
Montana        & 3    & 0    & 0    & 1    & 0    & 1    & 1    & 0    & 1    & 0    & 0    & 7     \\
Nebraska       & 0    & 0    & 1    & 1    & 0    & 0    & 0    & 6    & 1    & 0    & 4    & 13    \\
Nevada         & 0    & 1    & 1    & 0    & 3    & 1    & 7    & 2    & 1    & 0    & 1    & 17    \\
New hampshire  & 0    & 0    & 1    & 2    & 0    & 0    & 0    & 2    & 0    & 0    & 0    & 5     \\
New jersey     & 3    & 7    & 8    & 11   & 12   & 5    & 15   & 7    & 5    & 8    & 8    & 89    \\
New mexico     & 0    & 0    & 1    & 0    & 1    & 0    & 4    & 3    & 2    & 3    & 1    & 15    \\
New york       & 34   & 40   & 38   & 38   & 53   & 31   & 32   & 47   & 18   & 39   & 34   & 404   \\
North carolina & 1    & 0    & 13   & 19   & 8    & 9    & 5    & 9    & 4    & 7    & 8    & 83    \\
North dakota   & 0    & 1    & 0    & 0    & 0    & 0    & 0    & 0    & 0    & 0    & 1    & 2     \\
Ohio           & 5    & 9    & 14   & 13   & 12   & 11   & 8    & 10   & 9    & 8    & 16   & 115   \\
Oklahoma       & 7    & 15   & 9    & 18   & 22   & 8    & 8    & 6    & 2    & 3    & 5    & 103   \\
Oregon         & 3    & 1    & 2    & 4    & 12   & 3    & 3    & 14   & 2    & 2    & 1    & 47    \\
Pennsylvania   & 5    & 2    & 0    & 5    & 1    & 12   & 12   & 1    & 1    & 2    & 5    & 46    \\
Rhode island   & 0    & 0    & 2    & 0    & 0    & 1    & 0    & 0    & 0    & 0    & 0    & 3     \\
South carolina & 1    & 2    & 17   & 29   & 17   & 6    & 8    & 4    & 1    & 0    & 3    & 88    \\
Tennessee      & 0    & 1    & 2    & 0    & 5    & 7    & 10   & 2    & 5    & 2    & 18   & 52    \\
Texas          & 16   & 13   & 18   & 16   & 18   & 35   & 46   & 14   & 12   & 16   & 38   & 242   \\
Utah           & 0    & 2    & 0    & 1    & 0    & 0    & 0    & 2    & 1    & 0    & 0    & 6     \\
Vermont        & 0    & 1    & 0    & 0    & 0    & 0    & 1    & 1    & 0    & 0    & 2    & 5     \\
Virginia       & 1    & 3    & 6    & 2    & 16   & 6    & 5    & 4    & 9    & 5    & 4    & 61    \\
Washington     & 7    & 10   & 13   & 15   & 10   & 6    & 10   & 20   & 15   & 3    & 10   & 119   \\
West virginia  & 0    & 0    & 0    & 0    & 0    & 0    & 0    & 0    & 0    & 0    & 1    & 1     \\
Wisconsin      & 2    & 1    & 0    & 5    & 1    & 1    & 0    & 12   & 6    & 0    & 0    & 28    \\
Wyoming        & 0    & 0    & 0    & 0    & 0    & 0    & 1    & 0    & 0    & 0    & 0    & 1     \\ \hline
Total          & 223  & 397  & 435  & 498  & 510  & 468  & 478  & 551  & 352  & 300  & 454  & 4666  \\ \hline 
\end{tabular}
\caption{The number of articles about a crime occurring in a given state which contain
biased language in each year.}
\label{table:state_article_counts}
\end{table}

\end{document}